%% file: paper.tex
\documentclass[letterpaper, 10 pt, conference]{ieeeconf}  
\input{tex/preamble}

\providecommand{\sref}[1]{\S\ref{#1}}
\providecommand{\il}[1]{#1}
\providecommand{\figurename}{Fig.}

\renewcommand{\baselinestretch}{0.93} 

\begin{document}
\title{\LARGE \bf Learning Singularity Avoidance}
\author{Jeevan Manavalan$^{1*}$ \& Matthew Howard%
\thanks{%
$^{1}$Jeevan Manavalan and Matthew J. Howard are with the Centre for Robotics Research, Department of Informatics, King's College London, London, UK. \texttt{jeevan.manavalan@kcl.ac.uk}
}
\thanks{%
$^{*}$This work was partially supported by the EPSRC Grant Agreement No. EP/P010202/1.
}}
\maketitle
\thispagestyle{empty}
\pagestyle{empty}
\begin{abstract}
With the increase in complexity of robotic systems and the rise in non-expert users, it can be assumed that task constraints are not explicitly known. In tasks where avoiding singularity is critical to its success, this paper provides an approach, especially for non-expert users, for the system to learn the constraints contained in a set of demonstrations, such that they can be used to optimise an autonomous controller to avoid singularity, without having to explicitly know the task constraints. The proposed approach avoids singularity, and thereby unpredictable behaviour when carrying out a task, by maximising the learnt manipulability throughout the motion of the constrained system, and is not limited to kinematic systems. Its benefits are demonstrated through comparisons with other control policies which show that the constrained manipulability of a system learnt through demonstration can be used to avoid singularities in cases where these other policies would fail. In the absence of the systems manipulability subject to a tasks constraints, the proposed approach can be used instead to infer these with results showing errors less than $\mathbf{10^{-5}}$ in 3DOF simulated systems as well as $\mathbf{10^{-2}}$ using a 7DOF real world robotic system.
\end{abstract}

\section{Introduction}\label{s:introduction}
In recent years, there has been a booming shift from the development of specialised factory robots to versatile autonomous ones that are targeted at non-expert users. However, systems are performing tasks for which they were not specifically designed and there is uncertainty over the degree of redundancy (or contrastingly, overconstrainedness) in the control of their movements. Consequently, increasing importance is placed on improving control techniques to reduce such uncertainty, which otherwise may inadvertently affect the task at hand.

Traditionally, those issues are assessed by the so-called \textit{manipulability} of the system, by analysing the extent to which solutions exist to the inverse problem of finding control solutions for a given set of task constraints. First introduced by Yoshikawa \cite{yoshikawa1984analysis}, the manipulability index works by identifying linear dependencies in the task Jacobian that may cause a singular configuration to be reached. 

Initially used to devise control algorithms to avoid kinematic singularities in manipulators, it has since been used in a wide variety of contexts, such as real-time end-pose planning in walking tasks \cite{yang2016idrm}, grasp planning \cite{sundaram2016graspplanning}, and planning of human-robot interaction work spaces \cite{vahrenkamp2016interactionworkspace}. This measure has also been extended to account for joint limits, self-collision in redundant systems, and the need for adaptability to avoid obstacles in the work space \cite{vahrenkamp2012constrainedmanipulability}.

The common assumption among studies using different forms of Yoshikawa's manipulability index, \cite{Patel2015measureSurvey} is that the nature of the constraints affecting the system's manipulability (more specifically, the system's task Jacobian) \emph{is known analytically, \apriori} for design of the controller. However, with the rise in non-expert users and the increase in complexity of tasks, this assumption is increasingly untenable. Ignoring the manipulability in such systems risks, for instance, a non-expert user driving a robot through an unstable singular point and causing possible damage to the robot, or worse, injury to the user.

As an alternative, this paper provides a data-driven approach in which the constrained joint manipulability is \emph{learnt from user demonstrations}, without need for explicit definition through analytical approaches. This is beneficial to non-expert users without explicit knowledge of task constraints as it optimises a system's control to avoid instabilities in the system caused by singularities when carrying out a task. The proposed approach uses constraint consistent learning \cite{jeevan2017learningFast,learningspaceLin,dependantLearning2016,Howard2009} to, first, learn the task constraint and, second, optimise the manipulability derived from the learnt constraint matrix within the null space of the primary task constraint. It thereby, avoids singularity by maximising the \emph{learnt manipulability} throughout the motion of the constrained system, in the absence of analytical prior knowledge of the constraints. Results of estimating the manipulability subject to task constraints show errors less than $10^{-5}$ in 3DOF simulated systems as well as $10^{-2}$ using a 7DOF real world robotic system. 

\section{Problem Definition}\label{s:problem_definition}
\noindent This work considers the control of systems subject to uncertain constraints due to the complexity and/or naivety of non-expert users, and the need to prioritise joint configurations, which lead to greater degrees of freedom to flexibly perform demonstrated tasks.

\subsection{Task Prioritised Constraints} \label{s:taskPioritisedConstraints}
\noindent Formally, a system of $\dimb$-dimensional (self-imposed or environmental) constraints can be defined as

\begin{equation}
  \bA(\bx)\bu(\bx)=\bb(\bx)
  \label{e:problem-constraint}
\end{equation}
where $\bx\in\R^\dimx$ represents state (usually represented either in end-effector or joint space), $\bA(\bx)\in\R^{\dimb\times\dimu}$ is a matrix describing the constraints, $\bu\in\R^\dimu$ represents the action and   $\bb(\bx)\in\R^{\dimb}$ is the \emph{task space policy} describing the primary task to be accomplished. It is assumed that the $\bA(\bx)$ should be learnt by the robot through demonstrations, while also autonomously handling the degree of constrainedness of the system such that it enhances the flexibility.

The general solution to \eref{problem-constraint} is given by 
\begin{equation}
	\bu(\bx) = \underbrace{\bA^\dagger(\bx)\bb(\bx)}_{\buts}+ \underbrace{\bN(\bx)\bpi(\bx)}_{\buns}
\label{e:problem-solution}
\end{equation}
where $\buts$ is the task space component that implements the task space policy, $\buns$ is the null space component and 
\begin{equation}
	\bN(\bx):=\I-\bA(\bx)^\dagger\bA(\bx)\in\R^{\dimu\times\dimu}
	\label{e:N}
\end{equation}
is the null space projection matrix that projects the null space policy
$\bpi(\bx)$ onto the null space of $\bA$. Here, $\I\in\R^{\dimu\times\dimu}$
denotes the identity matrix and $\bA^\dagger=\bA^\T(\bA\bA^\T)^{-1}$ is the
Moore-Penrose pseudo-inverse of $\bA$. This does not only apply to kinematics, but also to redundant actuation \cite{tahara2009}, and redundancy in dynamics \cite{udwadia2007analytical}.

Commonly, in the context of programming by demonstration by non-expert users, $\bA, \bN, \bb$ and $\bpi$ are not explicitly known. Instead, the controller must be derived from data. In this paper, it is assumed that data is given as pairs of $\Nd$ observed states $\bx_n$ and actions $\bu_n$.

What non-expert users may not be aware of is that $\buns$ controls the additional free non-task related degrees of freedom of the system. It can be thought of as a lower priority task which does not conflict with the goal defined in the task space component. The benefit of having a null space component is evident in tasks which have multiple solutions. For example, a reaching task where multiple paths to a goal may be available, some paths may drive a system closer to its joint limits or singular configurations which can lead to an increased risk of  getting stuck or cause turbulent movement in face of perturbations or the imposition of additional constraints.

\subsection{Example: Opening Drawers}\label{s:hose_Attachment}
\begin{figure}[t!]
      \centering%
 \includegraphics[width=0.48\textwidth]{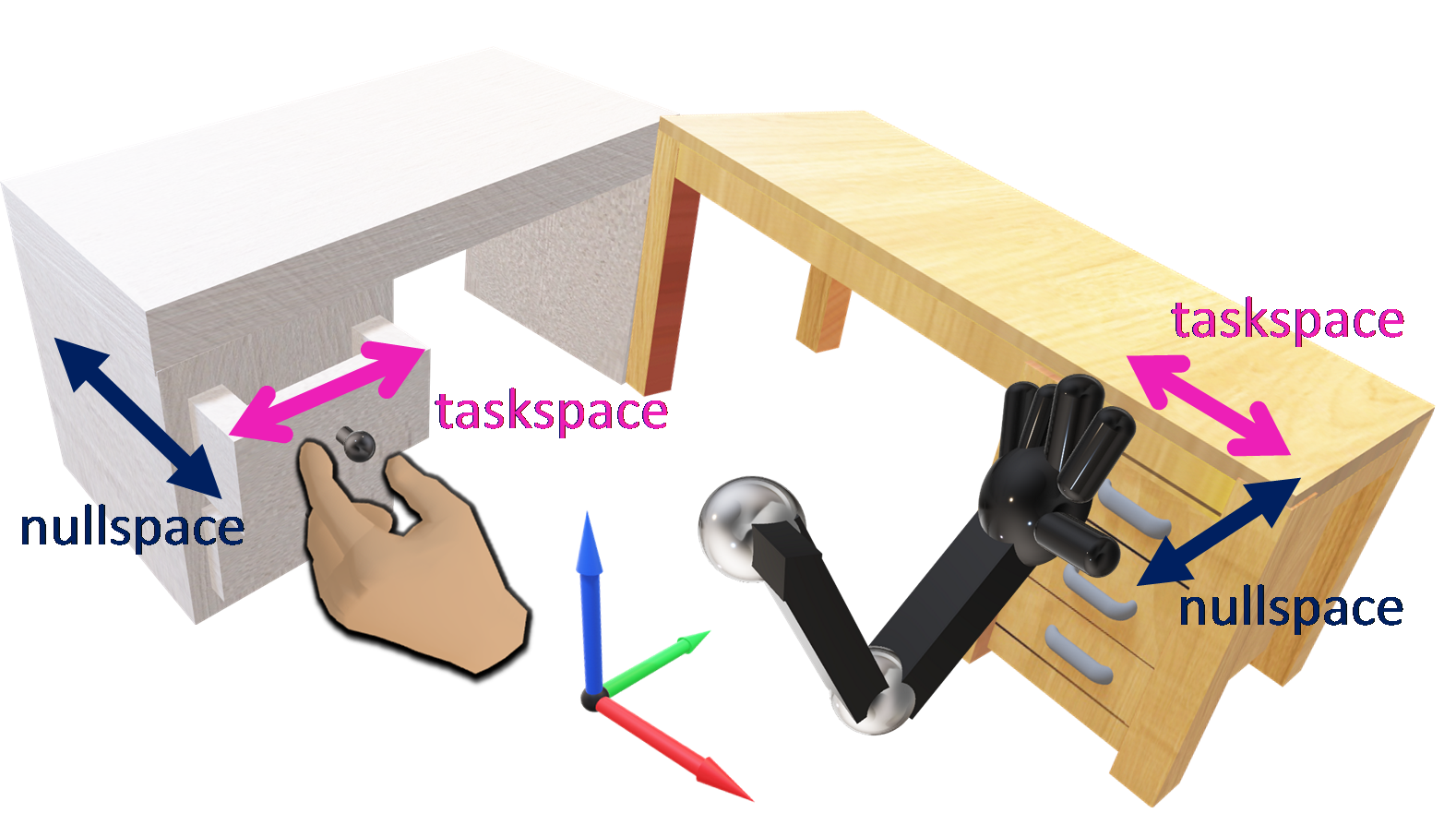}
      \caption{Demonstrating the task of opening drawers subject to their linear constraint.}
      \label{f:PbDRobotTask}
\end{figure}  
\noindent As a real world example, consider placing the robot in an environment designed for use by people such as an office. A robot trained to assist staff with, for example, collecting documents needs to be able to perform tasks such as opening filing cabinets. In this case, the problem of teaching the robot to open and close drawers can be solved through \PbD (as shown in \fref{PbDRobotTask}). Programming by demonstration is suitable because office staff can decide to retrain systems when required in another office or if the room layout changes without having any expert knowledge of the system and the need to call in a specialist. The primary task policy is to have the robot end-effector open the drawer, in which the constraint matrix $\bA$ is determined by a linear constraint following the direction of the drawer opening.

Where the task is kinaesthetically demonstrated, the user might manually guide a robot facing a filing cabinet, from holding onto a random point on the handle to opening the drawer towards the robot, and repeating this action multiple times using different starting points. In this scenario, the \textit{primary task} might be learnt in a straightforward manner using one of several constraint learning methods (see \sref{learningTheConstraint}).

Unless explicitly instructed, however,the user might not take specific care of how the motion appears in the null space when performing those demonstrations. For example, the user might choose to grab the drawer from the nearest point or in a pose/grip which allows demonstrations in the most comfortable manner for the user. It is unlikely that an average user will know to avoid unstable or singular configurations---however, these may occur in various task-dependant situations. If for example, a system placed in front of a drawer starts with all the joints fully extended, and a novice user then directs it to open the drawer by moving the arm directly back towards itself without resolving redundancy in the joint space, the system may propel itself in unpredictable directions due to the singularity. Moreover, systems that use an \textit{ad hoc} way of dealing with singularities, such as when using Matlab's pseudoinverse function which replaces singular values with zero are not satisfactory, as this prevents any movement of the system and thus comes at the expense of completing the task. A better way to do this is by maximising the system's manipulability.

\subsection{Task Manipulability and Programming by Demonstration}\label{s:manipulability-pbd}
\noindent To avoid these problems and reduce uncertainty of encountering unstable or singular configurations (\sref{hose_Attachment}), traditionally, Yoshikawa's manipulability index is used, whereby the null space degrees of freedom are used to maximise the distance from singular points during the execution of the primary task.

The manipulability is a measure of a system's ability to position and orientate its end-effector \cite{yoshikawa1984analysis}. In order to help with the designing and control of systems, Yoshikawa developed the manipulability index \cite{yoshikawa1984analysis}. It works by finding the linear dependencies in the task Jacobian which could result in reaching a singular configuration. Thus the following measure is used to assess manipulability
\begin{equation}
\v(\bx) = \sqrt{\det\left(\bA(\bx)\bA(\bx)^\T\right)} \label{e:manipulability-index-with-A}
\end{equation}
\noindent Note that, \textit{computation of the manipulability presupposes the availability of $\bA$ in analytical form}---however, as noted in \sref{taskPioritisedConstraints}, this is usually unavailable in the context of \PbD where the primary task and associated constraints are \emph{implicit in the demonstrations}.

As $\v\to 0$, the manipulability index indicates that the system is approaching a singular pose. The upper limit of $\v$ depends on the system itself and can only be provided once the entire work space is assessed (however the proposed method does not require this as it compares its manipulability locally). One of the applications of $\v$ lies in its use as a cost function to replace $\bpi$ in \eref{problem-solution} which results in the secondary task moving the system towards the goal using a path which favours higher manipulability (see \sref{constrainedV}).

To obtain $\v$, such that it can be used in such a manner, this paper proposes to \emph{form an estimate of the constraints in a task and derive an estimate of \eref{learningnullspacecomponent}}. The approach uses demonstration data such that it is applicable even for non-experts with no formal knowledge of the constraints involved in a task. In doing so, it allows for the robot to more accurately manoeuvre towards targets while minimising the risk of crossing over singular points and avoiding potentially unstable, unpredictable behaviour.

\section{Method}\label{s:method}
\noindent In this paper, the proposed approach forms an estimate of the task space constraint matrix $\bA$ to be used in $\bpi$ to manipulate systems away from singular points while performing a task. This minimises the risk of encountering singularities which lead to unpredictable behaviour.

\subsection{Data Collection}\label{s:representationOfData}
\noindent The proposed method works on data given as $\Nd$ pairs of observed states $\bx_n$ and actions $\bu_n$ collected though kinaesthetic demonstrations of the primary task. Assumptions on the data include that \il{\item observations are in the form presented in \eref{problem-solution}, \item $\bb$ from the task space varies throughout all observations, \item $\bpi$ from the null space is consistent throughout $\bu$, and \item $\bA$, $\bN$, $\bb$, and $\bpi$ are not explicitly known for any given observation}.

\subsection{Separating the task and null space component}\label{s:separatingNullspaceComponent}
\noindent Given the demonstration data $\{\bxn,\bun\}_{\nd=1}^\Nd$, the first step is to separate the task and null space components. For this, the approach first proposed in \cite{towell2010learning} can be used.

As shown there, the first and second terms of \eref{problem-solution} can be separated by seeking an estimate $\ebuns$ that minimises
\begin{equation}
  E[\ebuns] = ||\estimated{\boldsymbol{P}}_\nd\bun - \ebunsn||^2
  \label{e:learningnullspacecomponent}
\end{equation}

\noindent where $\ebunsn := \ebuns(\bxn)$ and $\estimated{\boldsymbol{P}}_\nd := {\ebunsn\ebunsn}^\T / ||\ebunsn||^2$. This works on the principal that, there exists a projection $\boldsymbol{P}$ for which $\boldsymbol{P}\bu = \boldsymbol{P}(\buts + \buns) = \buns$.

Similarly, $\ebuts$ is required as it functions as the primary task controller for the system and can be extracted by subtracting the newly estimated $\ebuns$ from $\bu$, \ie $\ebuts=\bu-\ebuns$.

\subsection{Representation \& Learning of the \noindent Constraint $\bA$}\label{s:learningTheConstraint}\label{s:RepresentationOfA}
\noindent At this point, the original demonstrated actions $\bu$ are separated into the task and null space parts. Based on the latter, the goal now is to compose an estimate of $\bA$ that can be used in assessing manipulability.
Constraints imposed on motion in the task space can refer to translational and orientational coordinates in the end-effector or joint space depending on the task at hand. An important criteria for using the manipulability for control is that constraints are \emph{state-dependant} \cite{lin2015learning}. Otherwise, if $\bA$ is constant across the state space \cite{dependantLearning2016} every state has the same manipulability index and the singularity avoidance controller has no role.

Taking this into consideration, a suitable representation of the constraint matrix is \cite{learningspaceLin}
\begin{equation}
	\bA(\bx) = \bLambda \bPhi(\bx)
	\label{e:A-matrix}
\end{equation}
where $\bLambda\in\R^{\dimb\times\dimPhi}$ is an unknown selection matrix (to be estimated in the learning) and $\bPhi(\bx)\in\R^{\dimPhi\times\dimu}$ is a feature matrix. The rows of the latter contain candidate constraints that can be predefined where there is prior knowledge of potential constraints affecting the system, or can take generic forms such as a series of polynomials. For example, one may choose $\bPhi(\bx)=\bJ(\bx)$, the Jacobian of the manipulator, where $\bA(\bx)=\bLambda\bJ(\bx)$ encodes constraints on the motion of specific degrees of freedom in the end-effector space.

Depending on the assumptions made on the representation of $\bA$, one of several learning methods could potentially be used to form the estimate of the selection matrix $\ebLambda$ \cite{dependantLearning2016,lin2015learning,armesto2017efficientlearning}. Of these, this paper picks \cite{lin2015learning} as it requires relatively few parameters and little data to perform robustly \cite{dependantLearning2016}. The estimate is formed by minimising
\begin{equation}
  E[\estimated{\bA}] =  \sum_{\nd=1}^\Nd \ebunsn^\T(\ebLambda\bPhi_\nd)^{\dagger}\ebLambda\bPhi_\nd\ebunsn
  \label{e:learninglambda}
\end{equation}
where $\bPhi_\nd:=\bPhi(\bxn)$. This results in the estimate $\ebA(\bx)=\ebLambda\bPhi(\bx)$.

\subsection{Estimated Singularity Avoidance Policy}\label{s:constrainedV}
\noindent Using the estimated constraint matrix $\ebA$, it is now possible to form the \textit{estimate of the system manipulability} for any configuration within the support of the data. The latter is given by substitution of $\bA$ in Yoshikawa's manipulability index.

\begin{equation}
  \ev(\bx) = \sqrt{\det\left(\ebA(\bx)\ebA(\bx)^\T\right)}.
  \label{e:manipulability-index-with-eA}
\end{equation}
%
\noindent From this constrained manipulability map, states for particular end-effector poses can be selected based on $\ev$ using $\bpi$ to update the joint angles. When used as a cost function, this information can provide the direction in which the system should move in order to increase its manipulability and maximise the distance from singular points, thereby reducing the risk of unpredictable behaviour. 
The simplest such approach is to use gradient ascent by replacing $\bpi$ in \eref{problem-solution} with
\begin{equation}
  \bpi_{\ev}(\bx) = \nabla_\bx\ev.
  \label{e:v cost function}
\end{equation}
Alternatively, if the task space trajectory is predictable, $\ev$ can be used in combination with global approximation in the null space (see, \eg \cite{nakamura1991advanced}).

\begin{figure}[t!]
    \vspace{1mm}
      \centering
      \framebox{\parbox{3in}{}
      \includegraphics[width=0.45\textwidth]{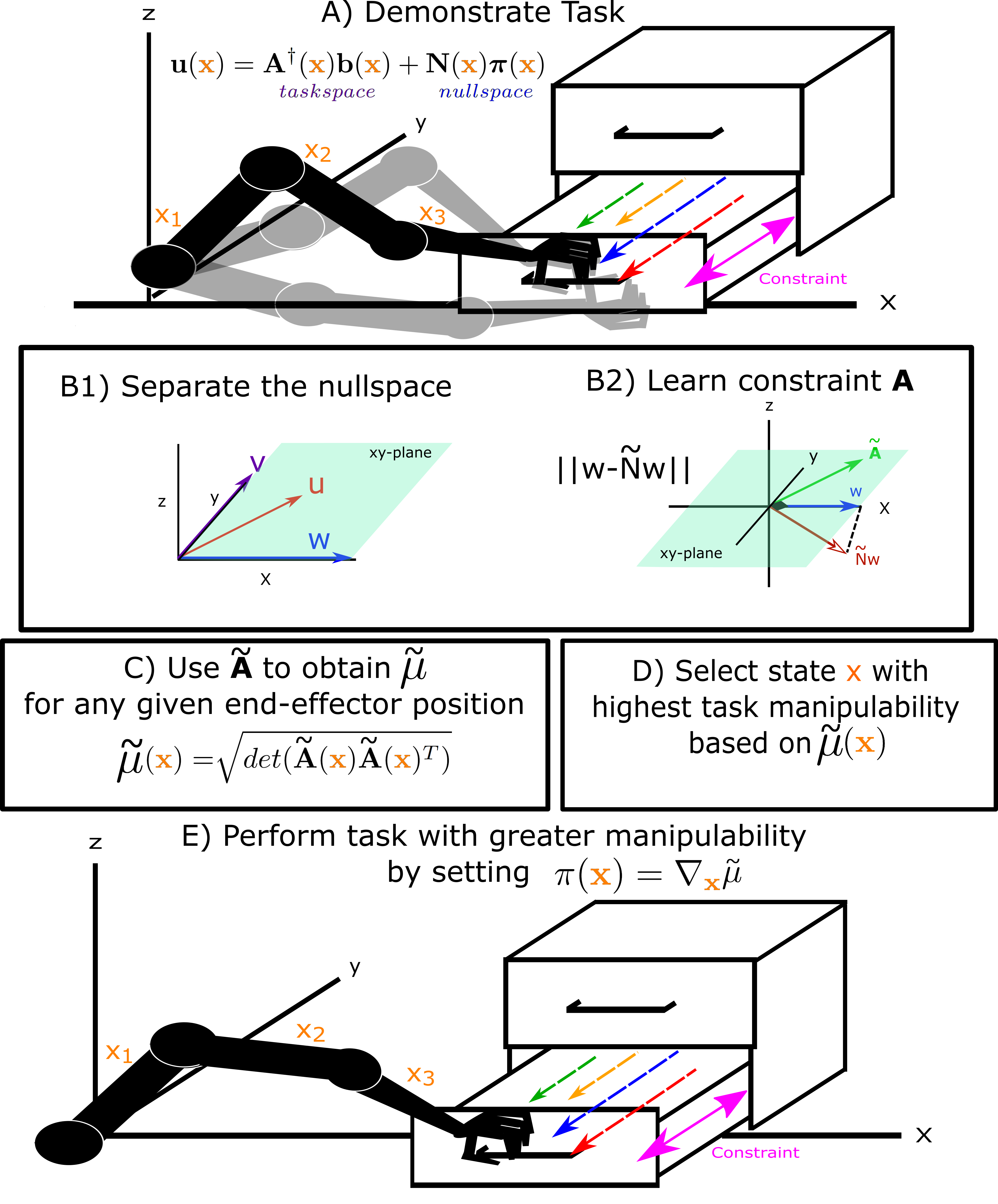}}
	  \caption{Overview of approach to maximising manipulability in \PbD tasks. (A) Motion data is collected through demonstrations of the task. (B1) The data is used to determine the separate task and null space components so that (B2) the latter can be used to estimate the constraint. (C) Using the estimated constraint matrix, an estimate of the constrained manipulability $\ev$ is made. (D) This estimate is used to select states with greater manipulability, and (E) control the robot toward these when performing the primary task.}
      \label{f:plan}
\end{figure}
\fref{plan} shows a brief overview of the major steps involved in the proposed approach.

 \subsection{Evaluation Criteria for Learning Methods}
 \noindent Learning the constrained manipulability may not always be a trivial matter depending on the task at hand. Factors such as, high dimensionality of a system, or the structure of particular constraints in comparison to others, can lead to poor learning performance. It is therefore necessary to define the following metric to assess learning performance.

 The Normalised Manipulability Index Error (NMIE) evaluates the fitness of the learnt manipulability index $\ev$, the following can be used to measure the distance between the true and learnt manipulability index

 \begin{equation}\label{NMIE}
 	E_{NMIE} =
 	\frac{1}{\mathcal{N}\sigma_\mathbf{\v}^2} \sum_{n=1}^{\mathcal{N}}||\mathbf{\v}_n - \mathbf{\ev}_n ||^2
 \end{equation}

\noindent where $\mathcal{N}$ is the number of data points. The error is normalised by the variance of $\mathbf{\v}$. The MIE will reach zero as $\mathbf{\ev}\to\mathbf{\v}$.



\section{Evaluation}\label{s:evaluation}
\noindent In this section, the proposed approach is first examined through a simulated 3-link planar system, before evaluating its performance in the context of \PbD of a physical robot.\footnote{The data supporting this research are openly available from King's College London at \texttt{http://doi.org/[link will be made available soon]}. Further information about the data and conditions of access can be found by emailing \texttt{research.data@kcl.ac.uk}} 

\subsection{Three Link Planar Arm}

\noindent The aim of the first evaluation is to test the robustness of learning the manipulability from motion data. The setup is as follows.

Constrained motion data is gathered from a kinematic simulation of a 3-link planar robot. The state and action space refer to the joint angle position and velocities, respectively, \ie $\bx:=\bq \in \mathbb{R}^3$ and $\bu:=\bqdot \in \mathbb{R}^3$. The taskspace is described by the end-effector coordinates $\mathbf{r} = (\r_x,\r_y,\r_\theta)^\T$ referring to the positions and orientation, respectively. The simulation runs at a rate of $50\,Hz$. 
Joint space motion of the system is recorded as it performs tasks under different constraints in the end-effector space. As described in \sref{RepresentationOfA}, a task constraint at state $\bx$ is described through
\begin{equation}
	\bA(\bx) = \bLambda \bJ(\bx)
	\label{e:AisLJ-matrix}
\end{equation}
where $\bJ \in \mathbb{R}^{3\times 3}$ is the manipulator Jacobian, and $\bLambda\in \mathbb{R}^{3 \times 3}$ is the selection matrix specifying the coordinates to be constrained. The following three constraints are evaluated:
\begin{align*}
	1) \ \ \bLambda_{x,y} = ((1,0,0),(0,1,0),(0,0,0))^T. \\
	2) \ \  \bLambda_{x,\theta} = ((1,0,0),(0,0,0),(0,0,1))^T. \\
	3) \ \  \bLambda_{y,\theta} = ((0,0,0),(0,1,0),(0,0,1))^T. \label{e:A-Constraint-Rows}
\end{align*}
\noindent To simulate demonstrations of reaching behaviour, the robot end-effector starts from a point chosen uniform-randomly $q_1  \sim U[0\degree,10\degree], q_2 \sim U[90\degree,100\degree], q_3 \sim U[0\degree,10\degree] $ to a taskspace target $\br^*$ and following a linear point attractor policy
\begin{equation}
  \bb(\bx) = \br^* - \br. 
\label{e:b-pointattractor}
\end{equation}
where $\br^*$ is drawn uniformly from $r^*_x \sim U[-1,1], r^*_y \sim U[0,2], r^*_{\theta} \sim U[0,\pi] $. Targets without a valid inverse kinematic solution are removed. All trajectories also use a point attractor as a control policy in $\buns$ 
\begin{equation}
  \bpi(\bx) = \pmb{\psi}^* - \bx.
\label{e:pi-pointattractor}
\end{equation}
The latter enforces consistency, that makes it easier to separate the constraint from the control policy. $\pmb{\psi}^*$ is arbitrarily chosen as $q_1=10\degree, q_2=-10\degree, q_3=10\degree$. For each constraint, $100$ trajectories are generated each containing 10 data points (1,000 points per constraint). This is done for separate training and testing data sets (a total of 2,000 points per constraint). Finally, this whole experiment is repeated $50$ times. 

\begin{table}[]
\vspace{4mm}
\captionof{table}{NMIE (mean$\pm$s.d.)$\times 10^{-3}$ for each constraint over 50 trials. \label{Tab:resultsTable}}
\centering
\begin{adjustbox}{width=8cm}
\begin{tabular}{lc}
\hline
\multicolumn{1}{c}{Constraint} & NMIE \\ \hline
$\bLambda_{x,y}$ & $ 0.001\times10^{-3} \pm 0.007\times10^{-3} $ \\ \hline
$\bLambda_{x,\theta}$ & $0.007\times10^{-3} \pm 0.004\times10^{-2} $ \\ \hline
$\bLambda_{y,\theta}$  & $0.002\times10^{-5}\pm 0.003\times10^{-5}$ \\ \hline
\end{tabular}
\end{adjustbox}
\vspace{-4mm}
\end{table}
The NMIE is presented in Table \ref{Tab:resultsTable}. As can be seen, learning of the constrained manipulability index is successful with errors less $10^{-5}$. This shows that the manipulability index can be learnt with very high precision through demonstrations, without having to explicitly know how the constraints affect the system's motions. 


To further assess the suitability of using $\ev$ instead of $\mu$, the RMSE is evaluated for 20 randomly generated trajectories of 100 points, using $\bpi_{\v}$ learnt under the constraint $\bLambda_{x,y}$ and $\bpi_{\ev}$ following \sref{constrainedV}. The starting point is chosen uniform-randomly $q_1  \sim U[0\degree,10\degree], q_2 \sim U[90\degree,100\degree], q_3 \sim U[0\degree,10\degree]$, and following \eref{b-pointattractor}, $r^*$ is drawn uniformly from $r^*_x \sim U[-1,1], r^*_y \sim U[0,2], r^*_{\theta} \sim U[0,\pi] $. This produces two trajectories, one using $\bpi_{\v}$ and the other $\bpi_{\ev}$. Results given as  (mean$\pm$s.d.)$\times 10^{-4}$ are $2.069 \pm 1.001$. These errors being lower than $10^{-3}$ in both the mean and standard deviation indicate that all 20 trajectories are accurately reproduced, therefore $\bpi_{\ev}$ is an appropriate replacement for when $\bpi_{\mu}$ is difficult to infer.

At this point, the suitability of $\bpi_{\ev}$ in absence of explicit knowledge of the constraint has been established. In order to understand the benefit of using the manipulability-based controller \eqref{e:v cost function}  to handle redundancy in the joint-space, its performance can be compared to that of other commonly used policies when encountering singularities in $\bLambda_{x,y}$. As examples of the latter, a zero policy and a linear point attractor in $\bpi$ are chosen. The zero policy emulates the most common and simplest approach (being the shortest path directly towards the target subject to the task constraints). The linear point attractor is also a common choice as a null space policy  as a way of bringing the arm to a default posture. When considering how systems behave near singular configurations, it is also important to consider the case where a system has an \textit{ad hoc} way to deal with singularities, which is becoming more common among safety protocols in commercialised systems for novice users (see \sref{manipulability-pbd}). Thus two cases are presented here, one where a system starts in a singular configuration without any impromptu way of dealing with singular values, and one case where singularities are dealt with by setting the singular value to zero after it crosses a certain threshold. Here, Matlab's \textit{ad hoc} approach is used, whereby the pseudoinverse function determines when singularities are encountered by using a threshold of \texttt{max(size(A)) $\times$ eps(norm(A))}, which in this case is $1.332\times 10^{-15}$. The \texttt{eps} in Matlab calculates the floating-point relative accuracy. 

\begin{figure}[t!]
\vspace{1mm}
      \centering
      \includegraphics[width=0.48\textwidth]{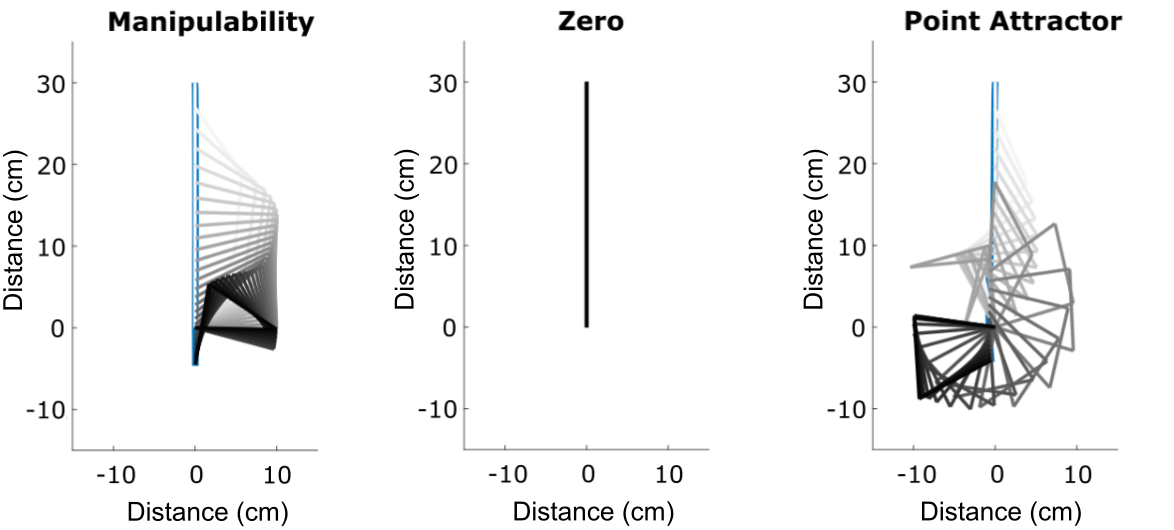}
     \includegraphics[width=0.48\textwidth]{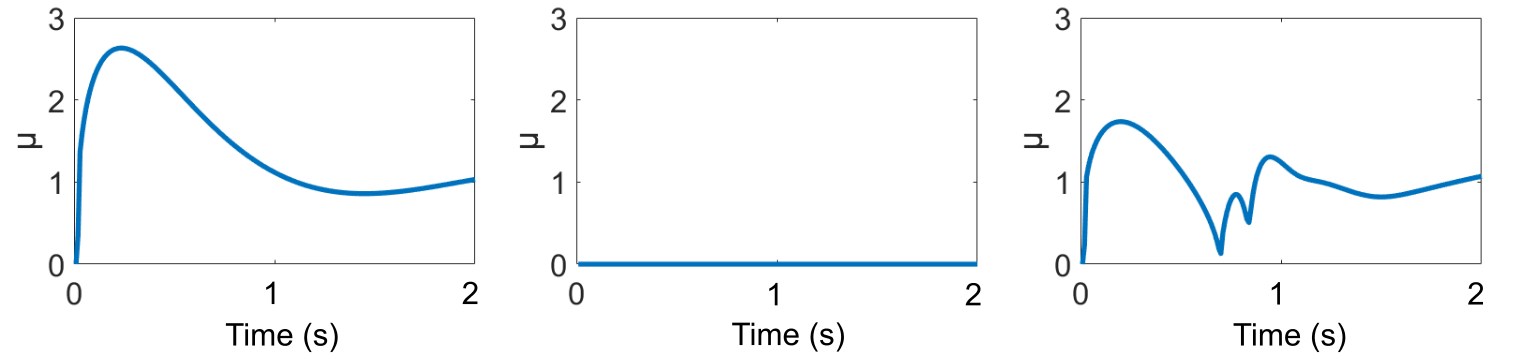}
      \caption{Comparing the manipulability, zero and point attractor in $\bpi$ where singular values in the taskspace are replaced with $0$. The bottom row shows the manipulability over time of the corresponding systems in the top row throughout the trajectory.}
      \label{compare1}
\end{figure}

Figure \ref{compare1} shows how a system behaves under the three different control policies in $\bpi$. This system is subject to a task constraint in the $r_x$ and $r_y$ coordinate and uses three control policies to evaluate how each handles movement from a singular pose. As shown, the proposed method is able to move away from the singular starting pose $q_1=(90 + 10^{-12})\degree$, $q_2=360\degree$ and $q_3=-360\degree$ using the learnt manipulability. On the other hand, the zero policy gets stuck at the starting point as it attempts to move directly down towards the target $\br^*=(0,0)$. The point attractor does succeed in the task, however, the movement to the default posture $q^*=(-190\degree,9\degree,-307\degree)^T$ results brings the robot close to singular configurations. While both the manipulability and point attractor policies reach their target, it is evident when looking at the bottom row that the (true) manipulability index of both systems are vastly different. As shown, the point attractor nearly approaches a singular configuration at around 70 steps into the movement which explains the overall erratic movement to reach the target. On the other hand, the manipulability encounters no such problem as it moves towards the target while maximising its manipulability throughout the movement. 

\begin{figure}[t!]
      \centering
      \includegraphics[width=0.48\textwidth]{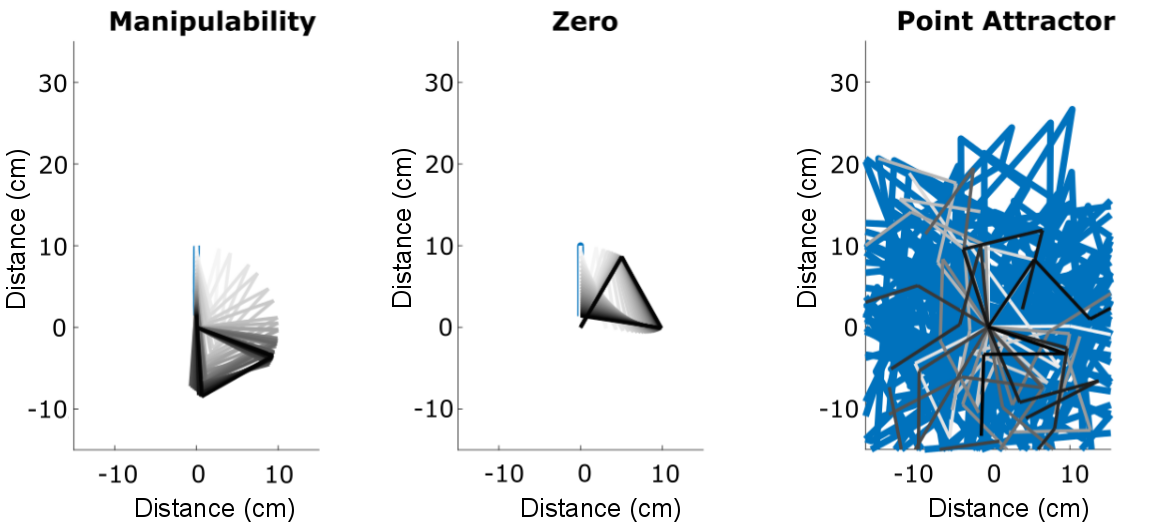}
     \includegraphics[width=0.48\textwidth]{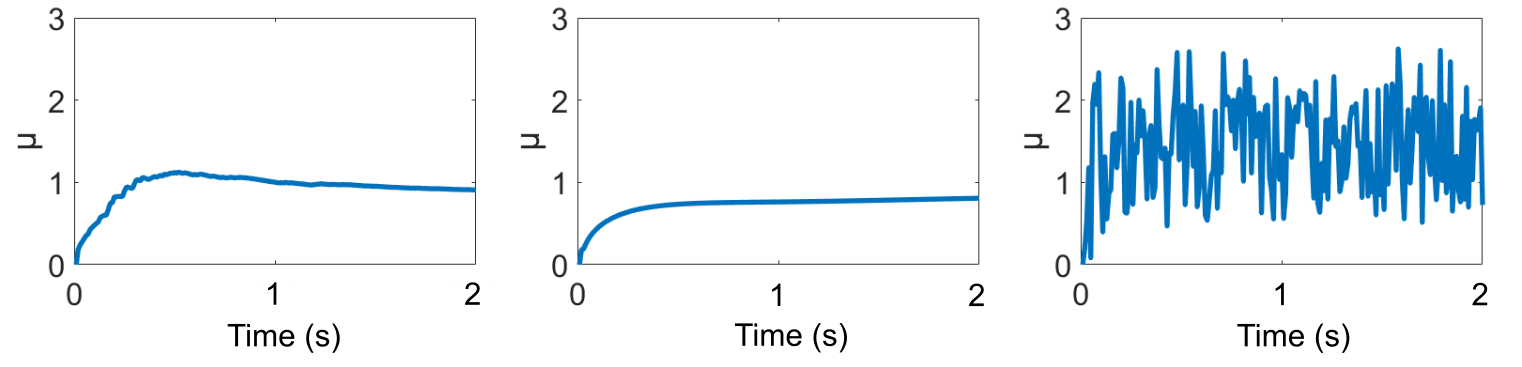}
     \caption{Comparing the manipulability, zero and point attractor in $\bpi$ where singularities are not dealt with. The bottom row shows the manipulability over time of the corresponding systems in the top row throughout the trajectory.}
     \vspace{-1mm}
      \label{compare2}
\end{figure}

Figure \ref{compare2} looks at a case where no \textit{ad hoc} method is used to detect singularities. In this case, a starting pose is set near a singular pose\footnote{as starting in a singular configuration would lead to a division over 0 in the taskspace within the first step of the system's movement regardless of the control policy.} of $q_1=90\degree$, $q_2=-180\degree$ and $q_3=(-180+ 10^{-10})\degree$. As shown, both the manipulaiblity and zero policy move away from the near-singular configuration. On the other hand, the point attractor with a  default posture  of $q^*=(-33\degree,-283\degree,193\degree)^T$  exhibits highly unstable behaviour (its next state exceeds $10^9$ for each joint). This type unpredictable behaviour is the most hazardous when having systems work in the real world.

Overall, these experiments show that the constrained manipulability of a system can be learnt through \PbD. Moreover, the learnt manipulability index can be used as a controller to avoid singularities in comparison to a straight-forward zero policy or a simple point attractor.



\subsection{7-Link Sawyer Arm}
\noindent The final experiment assesses the proposed approach in a real world task executed on the Sawyer, a 7DOF revolute physical robotic system with a maximum reach of 1260mm and precision of $\pm0.1$mm. The experimental scenario chosen is the closing of a drawer. 

The state and action space refer to the joint angle position and velocities, respectively, \ie $\bx:=\bq \in \mathbb{R}^7$ and $\bu:=\bqdot \in \mathbb{R}^7$. The taskspace is described by the end-effector coordinates $\br = (\r_x,\r_y,\r_z)^\T$ referring to positions in 3D cartesian space. The system runs at a rate of $100\,Hz$. 
Joint space motion is kinaesthetically recorded by guiding the Sawyer as it performs tasks under a constraint in the end-effector space. $\bJ \in \mathbb{R}^{3\times 7}$ is the manipulator Jacobian, and $\bLambda\in \mathbb{R}^{3 \times 3}$ is the selection matrix specifying the coordinates to be constrained. The following constraint is evaluated which models a drawer when it is orientated such that the constraint's null space lies along the $x$-axis (as shown in \fref{sawyerDemonstration})
\begin{align*}
	\ \ \bLambda_{x} = ((1,0,0),(0,0,0),(0,0,0))^T. 
 \label{e:A-Constraint-Rows}
\end{align*}

\noindent In order to have consistency in $\bpi$, the system starts in a default pose of  $q_1  \sim -100\degree, q_2 \sim 30\degree, q_3 \sim -100\degree, q_4 \sim 40\degree, q_5 \sim -60\degree, -70\degree, q_7 \sim 250\degree$ where the joints point outward such that stretching out the systems arm away from its body and along the constraint resolves redundancy in a similar manner for each trajectory.
\begin{figure}[t!]
\vspace{1mm}
     \centering
      \includegraphics[width=0.45\textwidth]{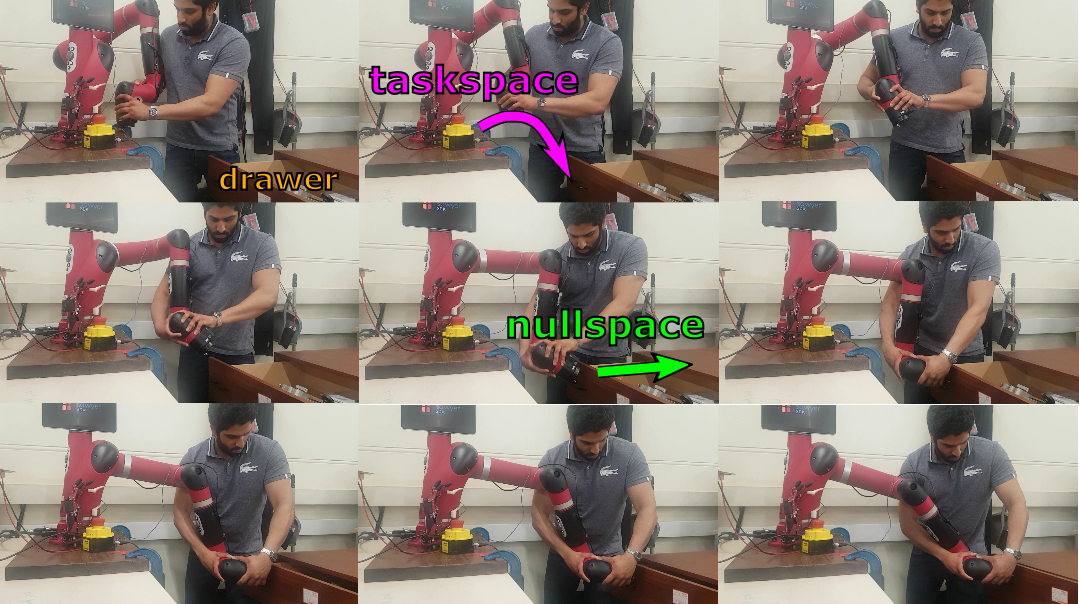}
      \caption{Learning the task constraint when closing the drawer through \PbD using the Sawyer.}
      \label{f:sawyerDemonstration}
      \vspace{-1mm}
\end{figure}

Fifty trajectories are recorded and the data is down-sampled such that each trajectory is reduced to 10 points. This is done as the direction of the constrained movements are captured even with such little data, and more data simply results in longer computation times.

The MIE learnt from 50 trajectories is $0.002$, therefore it is learnt successfully with errors below $10^{-2}$.\footnote{a similar experiment was done where the null space of the constraint lies along the $y$-axis giving alike performance} 
While it is shown that the manipulability index is learnt, it is important to establish whether the estimated manipulability is still suitable as a cost function to avoid singularity despite a greater error margin in comparison to the simulated 3DOF system. To this end, the RMSE is evaluated for 20 randomly generated trajectories of 100 points using the learnt model. The starting point is chosen uniform-randomly $q_1  \sim U[0\degree,10\degree], q_2 \sim U[90\degree,100\degree], q_3 \sim U[0\degree,10\degree], q_4 \sim U[90\degree,100\degree], q_5 \sim U[0\degree,10\degree], q_6 \sim U[90\degree,100\degree], q_7 \sim U[0\degree,10\degree]$. $r^*$ is drawn uniformly from $\sim U[-1,1]$ for the $x$-axis. Two trajectories are produced, one using $\bpi_{\v}$ and the other $\bpi_{\ev}$. The results are $0.220 \pm 0.094$ (mean$\pm$s.d.). Considering the high dimensionality of the robot, it is reasonable to assume an increased error in comparison to the simulated 3-link system.

\section{Conclusion}
\noindent This paper uses learning by demonstration to merge learning and manipulability-based control optimisation of an autonomous system to avoid singularities. The control optimisation uses a learnt cost function that maximises manipulability throughout the motion of a constrained system, not limited to kinematic systems. An approach is provided to learn the constraint of the task, if not known, from data. Results have been presented for a 3-link simulation in a 2D workspace and a real world experiment using the sawyer's arm in its 7DOF joint space. All experiments are in agreement that manipulabilities can be learnt through demonstration. The simulation demonstrates using the learnt manipulability as a cost function to have the system avoid singularities while performing a task. This approach is also verified in the real world using a robotic system with a high dimensional configuration space, showing that constraints can be learnt with enough precision to identify and avoid singular regions, when substituting $\ev$ for $\mu$ and being used as a cost function. When compared to other control policies such as a zero policy and a point attractor, the optimised movements from the proposed approach result in an autonomous system that moves towards the goal while handling redundancy by moving away from singular regions through local optimisation. This approach when compared to the aforementioned policies allows for the completion of the task, where the other policies are shown to succumb to the singularities within the same task resulting in either no movement at all or unpredictable behaviour.

Future work looks at conducting a study with naive subjects to evaluate the usefulness of the \PbD approach to avoid singularity in practice.

\addtolength{\textheight}{-12cm}   

\bibliographystyle{IEEEtran}





\end{document}

%% file: tex/preamble.tex
\IEEEoverridecommandlockouts                              
\usepackage{xcolor}
\usepackage{caption}
\usepackage{wrapfig}
\usepackage{graphics}
\usepackage[ruled,vlined]{algorithm2e}
\usepackage[noend]{algpseudocode}
\usepackage{multirow}
\usepackage{adjustbox}

\input{ext/symbols}

\input{ext/formatting}

\input{tex/abbreviations}
\input{tex/symbols}

\input{tex/comments}

%% file: ext/symbols.tex
\usepackage{amssymb}
\usepackage{mathtools}
\usepackage{amsmath}
\usepackage{gensymb}

\mathchardef\mhyphen="2D   

\newcommand{\R}     {\mathbb{R}}          
\newcommand{\T}     {\top}                
\newcommand{\I}     {\mathbf{I}}          

\newcommand{\estimated} [1]{\tilde{#1}}

\newcommand{\nd}      {n}                              
\newcommand{\Nd}      {\mathcal{\MakeUppercase{\nd}}}  

\newcommand{\bx}     {\mathbf{x}}         

\newcommand{\nx}     {p}                  
\newcommand{\dimx}   {\mathcal{\MakeUppercase{\nx}}} 

\newcommand{\bLambda}     {\mathbf{\Lambda}}         
\newcommand{\bA}     {\mathbf{A}}         
\newcommand{\bb}     {\mathbf{b}}  
\newcommand{\nb}     {s}                  
\newcommand{\dimb}   {\mathcal{\MakeUppercase{\nb}}} 
\newcommand{\bN}     {\mathbf{N}}         
\newcommand{\bw}     {\mathbf{w}}         
\newcommand{\bPhi}     {\mathbf{\Phi}}         
\newcommand{\bq}     {\mathbf{q}}         
\newcommand{\br}     {\mathbf{r}}         
\newcommand{\bJ}     {\mathbf{J}}         
\newcommand{\bqdot}  {\dot{\bq}}          

\newcommand  {\bu}  {\mathbf{u}}          
\renewcommand{\nu}  {q}                   
\newcommand  {\dimu}{\mathcal{\MakeUppercase{\nu}}} 
\newcommand  {\bpi} {\boldsymbol{\pi}}    

\newcommand{\bxn}    {\bx_{\nd}}          

\newcommand  {\ebA}     {\estimated{\bA}}  

%% file: ext/formatting.tex
\newcommand{\eg}{\textit{e.g.,}~} %
\newcommand{\ie}{\textit{i.e.,}~} %
\newcommand{\apriori}{\textit{a~priori}} %
\newcommand*{\sref}[1]{\S\ref{s:#1}}            
\newcommand*{\fref}[1]{\figurename~\ref{f:#1}}  
\newcommand*{\eref}[1]{(\ref{e:#1})}            

\let\labelindent\relax 
\usepackage[inline]{enumitem}
\setlist{nolistsep}
\newcommand{\il}[1]{\begin{enumerate*}[label=(\roman*)]#1\end{enumerate*}}

%% file: tex/abbreviations.tex
\input{ext/tex/abbreviations}



%% file: ext/tex/abbreviations.tex
\newcommand{\PbD    }{programming by demonstration\xspace}



%% file: tex/symbols.tex
\input{ext/tex/symbols}

\providecommand{\bxn}{\mathbf{x}_\nd}         
\providecommand{\bun}{\mathbf{u}_\nd}         
\providecommand{\ebunsn}{\ebuns_\nd}         

\providecommand{\buts }{\mathbf{v}}             
\providecommand{\ebuts}{\estimated{\buts}} 
\providecommand{\buns }{\bw}             
\providecommand{\ebuns}{\estimated{\buns}} 

\renewcommand  {\v    }{\mu}                    
\providecommand{\ev   }{\estimated{\v}}        

\providecommand{\ebLambda}{\estimated{\bLambda}}    

\renewcommand  {\r    }{r}                    

\providecommand{\dimPhi }{\mathcal{\MakeUppercase{p}}} 

%% file: ext/tex/symbols.tex
\usepackage{amssymb}
\usepackage{mathtools}

\mathchardef\mhyphen="2D   

\providecommand{\R}     {\mathbb{R}}          
\providecommand{\T}     {\top}                
\providecommand{\I}     {\mathbf{I}}          

\providecommand{\estimated} [1]{\tilde{#1}}

\providecommand{\nd}      {n}                              
\providecommand{\Nd}      {\mathcal{\MakeUppercase{\nd}}}  

\providecommand{\bx}     {\mathbf{x}}         
\providecommand{\nx}     {p}                  
\providecommand{\dimx}   {\mathcal{\MakeUppercase{\nx}}} 

\providecommand{\bA}     {\mathbf{A}}         
\providecommand{\bb}     {\mathbf{b}}         
\providecommand{\nb}     {s}                  
\providecommand{\dimb}   {\mathcal{\MakeUppercase{\nb}}} 
\providecommand{\bN}     {\mathbf{N}}         
\providecommand{\bw}     {\mathbf{w}}         
\providecommand{\bq}     {\mathbf{q}}         
\providecommand{\br}     {\mathbf{r}}         
\providecommand{\bJ}     {\mathbf{J}}         
\providecommand{\bqdot}  {\dot{\bq}}          

\providecommand  {\bu}  {\mathbf{u}}          
\providecommand  {\nu}  {q}                   
\providecommand  {\dimu}{\mathcal{\MakeUppercase{\nu}}} 
\providecommand  {\bpi} {\boldsymbol{\pi}}    

\providecommand{\bxn}    {\bx_{\nd}}          



%% file: tex/comments.tex
\usepackage[normalem]{ulem}                                        
\usepackage{marginnote}
\usepackage[textwidth=10ex,colorinlistoftodos]{todonotes}

\colorlet{jb}{red}
\colorlet{mh}{red}
\colorlet{yz}{blue}
\colorlet{jm}{green}
